\documentclass[10pt,twocolumn,letterpaper]{article}

\usepackage{cvpr}

\definecolor{cvprblue}{rgb}{0.21,0.49,0.74}
\usepackage[pagebackref,breaklinks,colorlinks,allcolors=cvprblue]{hyperref}

\usepackage[utf8]{inputenc} 
\usepackage[T1]{fontenc}    

\usepackage{url}            
\usepackage{booktabs}       
\usepackage{amsfonts}       
\usepackage{nicefrac}       
\usepackage{microtype}      
\usepackage{xcolor}         

\usepackage{algorithm}
\usepackage{algorithmic}
\usepackage{bbm}
\usepackage{bm}

\usepackage{wrapfig}

\usepackage{times}

\usepackage{graphicx}

\usepackage{amsmath}
\usepackage{amssymb}

\usepackage{multirow}

\usepackage[accsupp]{axessibility}  

\definecolor{RoyalBlue}{RGB}{65,105,225}
\definecolor{RedOrange}{RGB}{255,69,0}

\usepackage{makecell}
\usepackage{color, colortbl}
\definecolor{Gray}{gray}{0.9}
\definecolor{demphcolor}{RGB}{144,144,144}
\definecolor{airforceblue}{rgb}{0.36, 0.54, 0.66}

\usepackage{subcaption}

\def\paperID{15963} 
\def\confName{CVPR}
\def\confYear{2025}

\title{Scale Efficient Training for Large Datasets}

\author{
  Qing Zhou, Junyu Gao\thanks{Co-corresponding author.}, Qi Wang\footnotemark[1]\\
  Northwestern Polytechnical University \\
  {\tt\small\{chautsing, gjy3035, crabwq\}@gmail.com} \\
}

\begin{document}

\maketitle

\begin{abstract}
    The rapid growth of dataset scales has been a key driver in advancing deep learning research. However, as dataset scale increases, the training process becomes increasingly inefficient due to the presence of low-value samples, including excessive redundant samples, overly challenging samples, and inefficient easy samples that contribute little to model improvement.
    To address this challenge, we propose \textbf{S}cale \textbf{E}fficient \textbf{T}r\textbf{a}ining (SeTa) for large datasets, a dynamic sample pruning approach that losslessly reduces training time. To remove low-value samples, SeTa first performs random pruning to eliminate redundant samples, then clusters the remaining samples according to their learning difficulty measured by loss. Building upon this clustering, a sliding window strategy is employed to progressively remove both overly challenging and inefficient easy clusters following an easy-to-hard curriculum.
    We conduct extensive experiments on large-scale synthetic datasets, including ToCa, SS1M, and ST+MJ, each containing over 3 million samples.
    SeTa reduces training costs by up to 50\% while maintaining or improving performance, with minimal degradation even at 70\% cost reduction. Furthermore, experiments on various scale real datasets across various backbones (CNNs, Transformers, and Mambas) and diverse tasks (instruction tuning, multi-view stereo, geo-localization, composed image retrieval, referring image segmentation) demonstrate the powerful effectiveness and universality of our approach.
    Code is available at \url{https://github.com/mrazhou/SeTa}.
\end{abstract}
\section{Introduction}
Large-scale and high-quality datasets serve as the cornerstone of deep learning research - where data flows, intelligence grows. Carefully curated datasets like ImageNet~\cite{ImageNet} and COCO~\cite{MSCOCO1} have enabled the development of increasingly sophisticated models, which in turn drive the demand for even larger-scale datasets to fully leverage their learning capacity. This virtuous cycle between data and model scaling has consistently pushed the boundaries of deep learning capabilities~\cite{zhou2024larger,yu2024deep,hestness2017deep,liu2024datasets,resnet,vit}. Beyond manually annotated datasets, synthetic data generated through program rendering~\cite{ST,MJ}, web crawling~\cite{SS1M}, and Large Language Models (LLMs)~\cite{ToCa,touvron2023llama,llama2,jiang2023mistral} has emerged as a complementary approach to dataset creation, offering superior scalability and customization flexibility~\cite{lu2023machine}.
The availability of large-scale datasets, both manually curated and synthetically generated, has been pivotal in supporting effective model training. However, the computational efficiency of the training process itself has not kept pace with the growth in data volume. This discrepancy poses a significant challenge, particularly in resource-constrained environments where researchers must balance the benefits of extensive datasets against practical hardware limitations.
Therefore, developing efficient training strategies that can fully leverage large-scale datasets while maintaining practical computational requirements has become a key research focus~\cite{ucb,he2024large,InfoBatch,wang2024efficienttrain++,wang2023efficienttrain}.

\begin{figure}
    \centering
    
    \includegraphics[height=0.2\textheight]{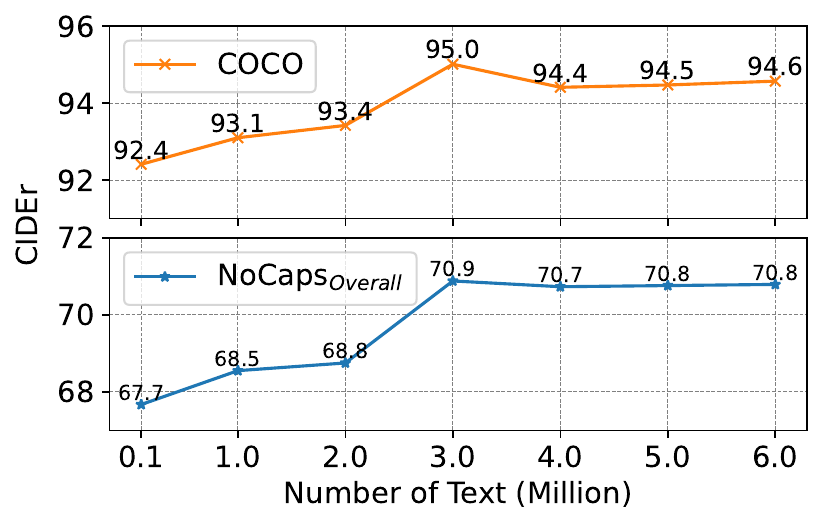}
    \vspace{-5pt}
    \caption{Image captioning performance of ViECap~\cite{ViECap} on COCO~\cite{COCO-cap} and NoCaps~\cite{NoCaps} tends to saturate as ToCa~\cite{ToCa} data volume increases above 3M.}
    \label{fig:mot}
    \vspace{-15pt}
\end{figure}

As the dataset size increases, the marginal benefit of additional redundant samples diminishes significantly, resulting in substantial computational overhead from ineffective training instances (as shown in Figure~\ref{fig:mot}).
Prior research has explored two main paradigms to address this challenge. Static coreset selection methods employ heuristic criteria~\cite{yang2022dataset,toneva2018empirical}, such as sample distances~\cite{xia2022moderate} and gradient magnitudes~\cite{paul2021deep}, to pre-filter redundant samples and construct a compact yet informative training subset. In contrast, dynamic loss-based pruning strategies~\cite{ucb,InfoBatch,wang2023efficienttrain} leverage sample-wise loss values as quality indicators, either through loss-based ranking or mean-loss comparison, to adaptively maintain a reduced high-value training set throughout the optimization process. Compared to static methods, efficiently implemented dynamic approaches~\cite{InfoBatch,ucb} offer several advantages: lower computational overhead as they avoid the expensive pre-processing step, better cross-architecture adaptability since they don't rely on model-specific features, and enhanced flexibility in handling diverse datasets as they continuously adapt to the model's learning progress.

However existing loss-based dynamic pruning methods effectively identify and exclude well-learned samples, they overlook other categories of low-value training instances, such as redundant duplicates and overly challenging samples that contribute minimally to model optimization despite consuming substantial computational resources. To address this limitation, we propose SeTa (Scale Efficient Training), a simple yet effective dynamic sample pruning framework that leverages loss-based difficulty estimation.
We utilize sample-wise loss as a computationally efficient proxy for learning difficulty, which naturally adapts to the model's evolving state during training. Our approach operates in two phases: First, we perform random sampling to eliminate redundant instances, followed by k-means clustering of the remaining samples based on their loss to create difficulty-stratified groups. Second, we employ a sliding window strategy that progressively shifts from easier to harder sample groups throughout the training process, effectively managing the curriculum of training samples.
The sliding process iterates multiple times based on the number of sample groups and training epochs. To ensure robust convergence, we incorporate an annealing mechanism in the final epochs, where a portion of the full dataset is randomly sampled to maintain training stability and reduce potential bias.

SeTa is a simple yet efficient training framework that can be seamlessly integrated into existing deep learning pipelines with only three lines of code modification~\cite{InfoBatch}. To demonstrate its effectiveness and versatility, we conduct extensive experiments across a diverse range of synthetic datasets, where the challenge of balancing data quantity and training efficiency is particularly pronounced. Specifically, we evaluate on three large-scale synthetic datasets: ToCa~\cite{ToCa} (3M samples generated via LLMs), SS1M~\cite{SS1M} (3M web-crawled samples), and ST+MJ~\cite{ST,MJ} (15M rendered samples). To further validate the generality of SeTa, we extend our evaluation to various architectures (ResNet, ViT, Swin, Vim and etc.) on the standard ImageNet benchmark, as well as across multiple domains including natural language processing (instruction tuning of LLMs), computer vision (multi-view stereo, geo-localization), and multimodal tasks (image captioning, composed image retrieval, referring image segmentation). Our comprehensive empirical study demonstrates that SeTa consistently achieves substantial training time reduction while maintaining or improving model performance, establishing its effectiveness as a general-purpose training acceleration framework.
\section{Related Work}

The majority of existing dynamic methods typically employ subsampling (selecting a portion of a dataset or a portion of a sample) to reduce training costs and achieve acceleration. Depending on the granularity of subsampling, methods can be classified into two categories: dataset and sample level.

Dataset-level methods aim to identify and train on the most informative subset of samples. Early works propose various importance metrics, such as gradient norm and error vector norm~\cite{paul2021deep}, or forgetting events~\cite{toneva2018empirical}, to select a core training set. While effective, these approaches typically require multiple training passes to identify the core set, introducing substantial computational overhead that can sometimes exceed the original training cost~\cite{InfoBatch}. Moreover, their metrics often show limited generalization across different datasets and model architectures.

Dynamic sampling methods address these limitations by continuously updating sample importance metrics during training. These approaches probabilistically exclude less informative samples while maintaining model performance. Dyn-Unc~\cite{he2024large} and UCB~\cite{ucb} introduces uncertainty-based metrics, while InfoBatch~\cite{InfoBatch} leverages readily available loss values as a proxy for sample importance. InfoBatch's use of loss values - a universal metric in deep learning - enables efficient pruning with minimal overhead and broad applicability. However, its strategy of only pruning samples below the mean loss potentially retains redundant and challenging training samples, particularly in large-scale datasets.

Complementary to dataset-level approaches, sample-level methods modify individual training instances. For instance, Tan~\textit{et al.}~\cite{tan2021efficientnetv2} employ downsampled images during early training phases, while EfficientTrain++~\cite{wang2023efficienttrain,wang2024efficienttrain++} introduces frequency-based image cropping and curriculum learning.
Despite their effectiveness, these image-centric approaches have limited cross-domain applicability and lack the plug-and-play convenience desired in deep learning frameworks.
\section{Method}
\subsection{Preliminaries}
Given a dataset $D = \{(x_1, y_1), ..., (x_n, y_n)\}$ where $x_i \in \mathcal{X} \subset \mathbb{R}^d$ represents the input and $y_i$ denotes the corresponding label. The objective of our training-efficient framework is to identify a sequence of minimal subsets $\{S_t\}_{t=1}^T$ across training epochs, where $S_t \subset D$ with $|S_t| < |D|$ for each epoch $t$. These subsets should ensure models achieve comparable performance to those trained on the full dataset $D$. Formally, we aim to solve:
\vspace{-5pt}
\begin{align}
    \begin{split}
        \min_{\{S_t\}_{t=1}^T} \sum_{t=1}^T |S_t|
        & \quad \text{s.t.} \quad |\mathcal{L}(\theta_{S_T}) - \mathcal{L}(\theta_D)| < \epsilon, \\
        \theta_{S_T} &= \arg\min_{\theta} \sum_{t=1}^T \mathbb{E}_{(x,y) \sim S_t}[\ell(\theta; x, y)],\\
        \theta_D &= \arg\min_{\theta} \mathbb{E}_{(x,y) \sim D}[\ell(\theta; x, y)],
    \end{split}
\end{align}
where $\theta_{S_T}$ represents the model parameters learned from the sequence of subsets after $T$ epochs, $\ell(\cdot)$ is the task-specific loss function, and $\epsilon$ is a small positive constant. The dynamic pruning ratio at epoch $t$ is defined as $\rho_t = 1 - |S_t|/|D|$, and the final pruning ratio across training is $\bar{\rho} = 1 - \frac{1}{T}\sum_{t=1}^T |S_t|/|D|$.
The saved training time is 
\begin{align}
\rho_O = \frac{|D| \times \bar{\rho} \times O_{m} + |D| \times O_{d}}{|D| \times O_{m}} = \bar{\rho} + \frac{O_{d}}{O_{m}},
\end{align}
where $O_{d, m}$ is single-sample time of data pruning and model training.
$\rho_O \approx \bar{\rho}$, when $O_{d} \ll O_{m}$.

\begin{figure}[t]
    \centering
    \includegraphics[width=1\linewidth]{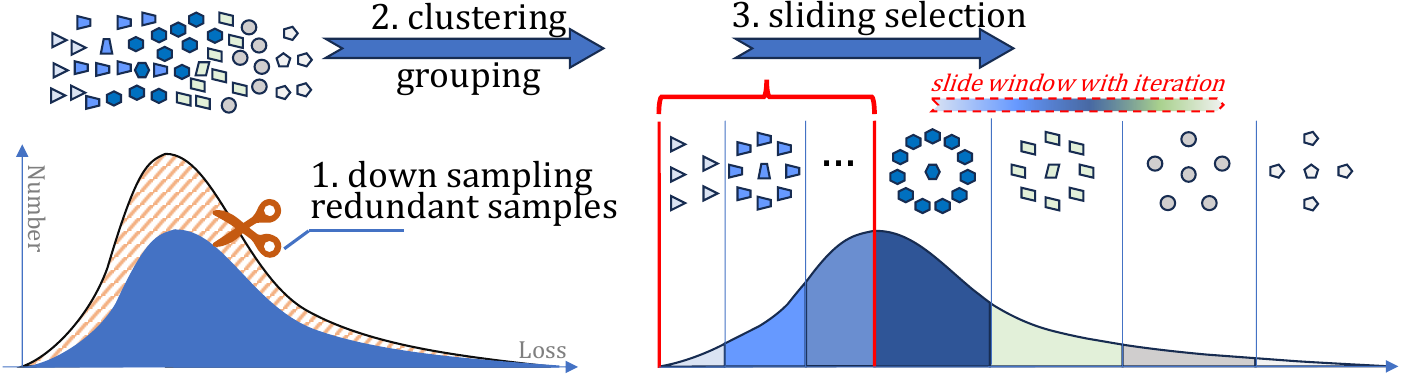}
    \caption{Overview of the proposed method for efficient training on large-scale datasets.}
    \label{fig:overall}
    \vspace{-15pt}
\end{figure}

\begin{table*}[ht]
    \caption{Comprehensive overview of 11 datasets, 10 tasks, and 14 models utilized in the experiments.}
    \label{tab:datasets}
    \centering
    \footnotesize
    \begin{tabular}{c|c|c|c|c|c|c}
        \toprule[1.3pt]
        Dataset & Type & Num. & Content & Source & Task & Model \\
        \midrule[1.3pt]
        Alpaca~\cite{alpaca} & \multirow{4}{*}{Syn.} & 52K & Instruction & LLM & Instruction Tuning & LLaMA-7B~\cite{touvron2023llama} \\
        ToCa~\cite{ToCa} &  & \textbf{3M} & Text & LLM & Zero-shot Visual Caption & ViECap~\cite{ViECap} \\
        SS1M~\cite{SS1M} &  & \textbf{3M} & Image, Text & Crawler & Zero-shot Image Caption & CapDec~\cite{CapDec} \\
        ST+MT~\cite{ST,MJ} &  & \textbf{15M} & Image, Text & Renderer & Scene Text Recognition & ABINet~\cite{ABINet} \\
        \hline
        CIRR~\cite{cirr} & \multirow{7}{*}{Real} & 22K & Image, Text & \multirow{7}{*}{\makecell{Human \\ annotated}} & Composed Image Retrieval & CaLa~\cite{jiang2024cala} \\
        WHU-MVS~\cite{WHU-MVS} &  & 28K & Image, Depth Map & & Multi-view Stereo & Ada-MVS~\cite{Ada-MVS} \\
        CVACT~\cite{CVACT} &  & 35K & Image & & Cross-view Geo-localization & GeoDTR~\cite{GeoDTR} \\
        CIFAR100~\cite{cifar100} & & 50K & Image & & Image Classification & ResNet~\cite{resnet} \\
        RefCOCO~\cite{refcoco} &  & 142K & Image, Text & & Referring Image Segmentation & CARIS~\cite{liu2023caris} \\
        COCO~\cite{COCO-cap} &  & 556K & Image, Text & & Image Caption & ViECap~\cite{ViECap} \\
        ImageNet~\cite{ImageNet} &  & \textbf{1.28M} & Image & & Image Classification & \makecell{ResNet~\cite{resnet}, ViT~\cite{vit}, Swin~\cite{liu2021swin}, Vim~\cite{vim}\\ MobileNetV3~\cite{mobilenetv3}, EfficientNet~\cite{tan2019efficientnet}} \\
        \bottomrule[1.3pt]
    \end{tabular}
\end{table*}

\subsection{Efficient Training}
We propose SeTa, a simple framework for efficient training through dynamic sample selection as shown in Figure~\ref{fig:overall}. At its core, SeTa employs sample-wise loss as an intrinsic difficulty measure, which serves as a model-agnostic and computation-free proxy for learning dynamics.

\noindent{\textbf{Loss-guided Sample Clustering.}}
To handle the redundancy in large-scale datasets, we first perform uniform down-sampling with ratio $r \in (0, 1)$, yielding a subset $\mathcal{I} = \{i_1, ..., i_m\}$ where $m = r|D|$ and $i_j \sim Uniform(1, |D|)$. The sampled instances are then partitioned into $k$ clusters through loss-guided k-means optimization:
\begin{align}
    \mathcal{C}^* = \arg\min_{\mathcal{C}} \sum_{j=1}^k \sum_{x_i \in \mathcal{G}_j} \|l_i^t - c_j\|^2,
\end{align}
where $\mathcal{C} = \{c_1, ..., c_k\}$ denotes the cluster centroids and $\mathcal{G}_j$ represents the sample set in cluster $j$, $l_i^t$ denote the loss of sample $i$ at training step $t$. The optimization iterates through:
\begin{align}
    & group(x_i) = \arg\min_{j} \|l_i^t - c_j\|^2, \\
    & c_j = \frac{1}{|\mathcal{G}_j|} \sum_{x_i \in \mathcal{G}_j} l_i^t.
\end{align}

Upon convergence, we obtain difficulty-stratified clusters $\{\mathcal{G}_1, ..., \mathcal{G}_k\}$ ordered by their centroid values.

\noindent{\textbf{Sliding Window Selection.}}
To orchestrate optimal learning progression, we introduce a dynamic curriculum mechanism that cyclically exposes the model to samples of increasing difficulty. We establish a window size $w = \lceil \alpha k \rceil$, where $\alpha \in (0, 1]$ controls the proportion of clusters selected at each iteration. The window position evolves according to:
\begin{align}
    & s_t = n \text{ mod } (k - w + 1), \\
    & e_t = s_t + w - 1.
\end{align}
where $n$ represents the current iteration count. This formulation ensures that when $s_t + w$ exceeds $k$, the window position resets to 0, implementing a cyclical curriculum that repeatedly progresses from easy to difficult samples. The selected sample subset is then constructed as:
$\mathcal{S}_t = \bigcup_{j=s_t}^{e_t} \mathcal{G}_{\sigma(j)}$,
where $\sigma(\cdot)$ represents the permutation that sorts clusters by their difficulty. This cyclic mechanism ensures that the model regularly revisits easier samples while progressively adapting to more challenging ones, maintaining stable optimization dynamics throughout the training process.

\noindent{\textbf{Partial Annealing.}}
To mitigate potential optimization bias from localized sample selection, we introduce a stochastic partial annealing strategy during the final training phase. Unlike conventional full-dataset annealing~\cite{InfoBatch}, we maintain efficiency by applying probabilistic sample selection with a fixed sampling ratio $r$:
\begin{align}
    \mathcal{S}_t^{anneal} = \{x_i | x_i \in \mathcal{G}, u_i < r\},
\end{align}
where $u_i \sim Uniform(0,1)$ and $\mathcal{G} = \cup_{j=1}^k \mathcal{G}_j$. This straightforward yet effective approach achieves comparable convergence to full annealing while maintaining computational efficiency through controlled sample exposure.

\section{Experiments}
\subsection{Datasets and Settings}
\newcommand{\blue}[1]{$_{\color{RoyalBlue}\downarrow #1}$}
\newcommand{\mred}[1]{$_{\color{red}\uparrow #1}$}
\textbf{Datasets.} We conduct experiments on synthetic datasets obtained through three representative large-scale synthetic data acquisition methods, including: 1) ToCa, which consists of 3 million text samples generated based on LLM; 2) SS1M, comprising 3 million samples obtained through web crawling; 3) ST+MT, containing 15 million image-text pairs rendered using a program engine. In addition to these large-scale synthetic datasets, we also carry out extensive experiments on large-scale real datasets such as ImageNet and COCO, as well as small-scale datasets including CIRR, WHU-MVS, CIFAR100, RefCOCO, and CVACT, covering various tasks and different models. For detailed information regarding  datasets, tasks, and models, please refer to Table~\ref{tab:datasets}.

\noindent{\textbf{Settings.}} To ensure the fairness and reproducibility of the experimental results, we use the corresponding default optimal settings for different tasks, models, and datasets.
This also means that SeTa can be seamlessly integrated into the existing data training process without adjusting any parameters in the existing process.
To demonstrate the efficiency, lossless, and acceleration of SeTa, we conduct three experiments with different settings for each task: L1 - improving performance with low pruning rate; L2 - comparable performance with high pruning rate; L3 - slightly degrading performance with extremely high pruning rate.
Pruning ratio offers a hardware-agnostic metric for computational efficiency, avoiding runtime noise from system variability. SeTa’s lightweight data selection (negligible vs. training time, satisfy Eq. 2) ensures $\rho_O \approx \bar{\rho}$. Therefore, we primarily report Pruned \% ($\bar{\rho}$).

\begin{table*}[h]
    \centering
    \caption{Comparison of state-of-the-art methods on CIFAR10 and CIFAR100 with ResNet18 at different pruning rates. $^{\dag}$: adjusted epochs in original paper. $^{\ddagger}$: same number of epochs as other methods. \textbackslash{}: unattainable rate. \textit{Italics}: reproduced results.}
    \label{tab:sota}
    \footnotesize
    \begin{tabular}{c|ccc|ccc}
        \toprule[1.3pt]
        \multirow{2}{*}{Method} & \multicolumn{3}{c|}{CIFAR10} & \multicolumn{3}{c}{CIFAR100} \\
        & 30\% & 50\% & 70\% & 30\% & 50\% & 70\% \\
        \midrule[1.3pt]
        ResNet18 & \multicolumn{3}{c|}{95.6} & \multicolumn{3}{c}{78.2 / \textit{79.3}} \\
        \hline
        Static Random & 94.6 \blue{1.0} & 93.3 \blue{2.3} & 90.2 \blue{5.4} 
            & 73.8 \blue{4.4} & 72.1 \blue{6.1} & 69.7 \blue{8.5} \\
        Dynamic Random & 94.8 \blue{0.8} & 94.5 \blue{1.1} & 93.0 \blue{2.6} 
            & 77.3 \blue{0.9} & 75.3 \blue{2.9} & - \\
        $\epsilon$-greedy~\cite{ucb} & 95.2 \blue{0.4} & 94.9 \blue{0.7} & 94.1 \blue{1.5}
            & 76.4 \blue{1.8}& 74.8 \blue{3.4} & - \\
        UCB~\cite{ucb} & 95.3 \blue{0.3} & 94.7 \blue{0.9} & 93.9 \blue{1.7} 
        & 77.3 \blue{0.9} & 75.3 \blue{2.9} & - \\

        InfoBatch$^{\dag}$~\cite{InfoBatch} & \colorbox{lightgray}{95.6 \mred{0.0}} & 95.1 \blue{0.5} &  94.7 \blue{0.9} 
            & \colorbox{lightgray}{78.2 \mred{0.0}}& 78.1 \blue{0.1} & 76.5 \blue{1.7} \\
        \hline
        InfoBatch$^{\ddagger}$~\cite{InfoBatch} & 95.3 \blue{0.3} & 95.0 \blue{0.6} & 94.3 \blue{1.3}
            & \textit{79.2 \blue{0.1}} & \textit{78.2 \blue{1.1}} & \textbackslash \\
        SeTa (ours) & \colorbox{lightgray}{\textbf{95.7} \mred{0.1}} & \textbf{95.3} \blue{0.3} & \textbf{95.0} \blue{0.6}
            & \textit{\colorbox{lightgray}{\textbf{79.4} \mred{0.1}}} & \textit{\textbf{79.0} \blue{0.3}} & \textit{\textbf{77.7} \blue{1.6}} \\
        \bottomrule[1.3pt]
    \end{tabular}
\end{table*}

\begin{table}[ht]
    \caption{ToCa including 3M text samples synthesized by LLM for zero-shot visual captioning task.}
    \label{tab:toca}
    \centering
    \footnotesize
    \setlength{\tabcolsep}{-0.5pt}
    \begin{tabular}{cc|cccc|c}
        \toprule[1.3pt]
        \multirow{2}{*}{Method} & \multirow{2}{*}{Pruned \%} & \multicolumn{4}{c|}{NoCaps Val (CIDEr)~\cite{NoCaps}} & \multicolumn{1}{c}{COCO} \\
        ~ & ~ & In & Near & Out & Overall & Test \\
        \midrule[1.3pt]
        \multirow{1}{*}{ViECap} & - 
            & 63.2 & 68.8 & 70.2 & 70.5 & 95.2 \\
        \hline

        \multirow{3}{*}{+InfoBatch} & 23.6
            & \colorbox{lightgray} {63.4 \mred{0.2}}    & 68.2 \blue{0.6}    & \colorbox{lightgray} {70.4 \mred{0.2}}    & 70.2 \blue{0.3} & 94.4 \blue{0.8} \\
            
            ~ & 34.1
            & 61.9 \blue{1.3}    & 67.1 \blue{1.7}    & 69.6 \blue{0.6}    & 69.2 \blue{1.3} & 93.7 \blue{1.5} \\
            
            ~ & 45.9
            & 60.8 \blue{2.4}    & 65.7 \blue{3.1}    & 67.5 \blue{2.8}    & 67.7 \blue{2.8} & 93.0 \blue{2.3} \\
        \hline

        \multirow{3}{*}{+SeTa} & 31.7
            & \colorbox{lightgray} {63.4 \mred{0.2}}    & \colorbox{lightgray} {69.7 \mred{0.9}}    & \colorbox{lightgray} {71.4 \mred{1.2}}    & \colorbox{lightgray} {71.5 \mred{1.0}} & \colorbox{lightgray} {95.3 \mred{0.1}} \\
            
            ~ & 41.8
            & \colorbox{lightgray} {63.9 \mred{0.7}}   & 67.2 \blue{1.6}    & 69.5 \blue{0.7}    & 69.3 \blue{1.2} & 94.5 \blue{0.7} \\

            ~ & 50.1
            & 61.5 \blue{1.7}    & 67.1 \blue{1.7}    & 69.9 \blue{0.3}    & 69.4 \blue{1.1} & 94.6 \blue{0.6} \\
        \bottomrule[1.3pt]
    \end{tabular}
\end{table}

\begin{table}[ht]
    \caption{SS1M including 3M text samples obtained through web crawling for zero-shot and cross-domain image captioning task.}
    \label{tab:ss1m}
    \centering
    \footnotesize
    \setlength{\tabcolsep}{0pt}
    \begin{tabular}{cc|cc|cc}
        \toprule[1.3pt]
        \multirow{2}{*}{Method} & \multirow{2}{*}{Pruned \%} & \multicolumn{2}{c|}{COCO} & \multicolumn{2}{c}{Flickr30k~\cite{Flickr30k}} \\
        ~ & ~ & B@4 & C & B@4 & C \\
        \midrule[1.3pt]
        \multirow{1}{*}{CapDec} & - 
            & 9.3 & 42.4
            & 7.7 & 23.1
            \\
        
        \hline

        \multirow{3}{*}{+InfoBatch} & 21.9
            & 9.1 \blue{0.2} & 41.9 \blue{0.5}
            & \colorbox{lightgray} {8.3 \mred{0.6}} & \colorbox{lightgray} {23.9 \mred{0.8}}
            \\
            
            ~ & 31.3
            & \colorbox{lightgray} {9.4 \mred{0.1}} & 42.1 \blue{0.3}
            & \colorbox{lightgray} {7.9 \mred{0.2}} & \colorbox{lightgray} {23.1 \mred{0.0}}

            \\
            
            ~ & 41.9
            & \colorbox{lightgray} {9.5 \mred{0.2}} & 42.3 \blue{0.1}
            & 7.6 \blue{0.1} & 21.3 \blue{1.8}
            \\
        
        \hline

        \multirow{3}{*}{+SeTa} & 40.0
            & \colorbox{lightgray} {10.3 \mred{1.0}} & \colorbox{lightgray} {45.1 \mred{2.7}}
            
            & \colorbox{lightgray} {9.2 \mred{1.5}} & \colorbox{lightgray} {24.8 \mred{1.7}}
            \\
            
            ~ & 51.6
            & \colorbox{lightgray} {9.9 \mred{0.5}} & \colorbox{lightgray} {44.0 \mred{1.6}}
            
            & \colorbox{lightgray} {8.0 \mred{0.3}} & \colorbox{lightgray} {23.1 \mred{0.0}}
            \\
            
            ~ & 62.1
            & 9.2 \blue{0.1} & 42.1 \blue{0.3}
            
            & 7.5 \blue{0.2} & 21.9 \blue{1.2}
            \\
        
        \bottomrule[1.3pt]
    \end{tabular}
\end{table}

\begin{table}[ht]
    \caption{MJ+ST including 15M image-text samples rendered by engine for scene text recognition task.}
    \label{tab:str}
    \centering
    \footnotesize
    \setlength{\tabcolsep}{-0.5pt}
    \begin{tabular}{cc|ccccc}
        \toprule[1.3pt]
        Method & ~Pruned \% & IIIT5k & SVT & IC15 & SVTP & CUTE80 \\
        \midrule[1.3pt]
        ABINet & - & 96.1 & 93.4  & 85.4 & 88.7 & 89.2 \\
        \hline

        \multirow{3}{*}{+InfoBatch}
            & 26.6
            & \colorbox{lightgray} {96.3 \mred{0.2}} & \colorbox{lightgray} {93.6 \mred{0.2}} & \colorbox{lightgray} {85.4 \mred{0.0}} & \colorbox{lightgray} {88.8 \mred{0.1}} & \colorbox{lightgray} {89.4 \mred{0.2}} \\

            & 38.1
            & 95.9 \blue{0.2} & 93.2 \blue{0.2} & 84.2 \blue{1.2} & 88.0 \blue{0.7} & 88.4 \blue{0.8} \\

            & 50.3
            & 95.8 \blue{0.3} & 93.2 \blue{0.2} & 84.0 \blue{1.4} & 88.1 \blue{0.6} & 88.5 \blue{0.7} \\
        \hline

        \multirow{3}{*}{+SeTa}
            & 28.1
            & \colorbox{lightgray} {96.2 \mred{0.1}} & \colorbox{lightgray} {93.4 \mred{0.0}} & \colorbox{lightgray} {85.6 \mred{0.2}} & \colorbox{lightgray} {89.3 \mred{0.6}} & \colorbox{lightgray} {89.2 \mred{0.0}} \\

            & 40.4
            & 95.9 \blue{0.2} & 93.2 \blue{0.2} & \colorbox{lightgray} {85.4 \mred{0.0}} & 88.0 \blue{0.7} & 88.5 \blue{0.7} \\

            & 71.0
            & 95.8 \blue{0.3} & 92.9 \blue{0.5} & 84.9 \blue{0.5} & 88.2 \blue{0.5} & 87.5 \blue{1.7} \\
        \bottomrule[1.3pt]
    \end{tabular}
\end{table}
\subsection{Efficiency Evaluation and Comparison}
We conduct comprehensive experiments (Table~\ref{tab:sota}, \ref{tab:toca}, \ref{tab:ss1m}, \ref{tab:str}) on large-scale datasets to evaluate SeTa's efficiency and compare it with state-of-the-art methods.

\noindent{\textbf{Comparison with SOTA Methods.}} As shown in Table~\ref{tab:sota}, SeTa achieves consistent improvements over state-of-the-art methods on both CIFAR10 and CIFAR100 across pruning rates. Notably, at the extreme 70\% pruning rate, SeTa attains 95.0 on CIFAR10 (merely -0.6 drop from baseline) and 77.7 on CIFAR100 (-1.6 drop), while InfoBatch$^\ddagger$ fails to reach this pruning rate (\textbackslash). In our reproduced experiments with matched training epochs ($^{\ddagger}$), SeTa surpasses InfoBatch$^{\ddagger}$ on CIFAR100 at 50\% pruning. The method also surpasses other approaches like UCB by +1.1 points on CIFAR10 under 70\% pruning. Crucially, SeTa achieves 95.7 at 30\% pruning on CIFAR10 (surpassing the baseline 95.6) and 79.4 on CIFAR100 (exceeding the baseline \textit{79.3}), demonstrating both stability and accuracy improvements. These results validate SeTa’s effectiveness in maintaining performance under aggressive pruning without requiring epoch adjustments.

\noindent{\textbf{Superior Performance-Preservation Trade-off (L1).}} At low pruning rates designed to improve performance, SeTa demonstrates remarkable ability to maintain or even enhance model performance while achieving substantial data reduction. On ToCa dataset, SeTa achieves a 1.0-point improvement in Overall CIDEr (71.5 vs 70.5) while pruning 31.7\% of data, whereas InfoBatch exhibits performance degradation even with a lower 23.6\% pruning rate. This superior trade-off is consistently observed across datasets, particularly evident in SS1M where SeTa's pruned dataset yields significant improvements across all evaluation metrics (e.g., B@4 gains of 1.0 and 1.5 on COCO and Flickr30k). It demonstrating SeTa's enhanced ability to progressively remove low-value samples while maintaining the core learning curriculum.

\noindent{\textbf{Enhanced Robustness at Higher Pruning Rates (L2).}} In the moderate-to-high pruning regime designed to maintain performance, SeTa exhibits exceptional robustness in preserving model performance. Notably, on ToCa, SeTa maintains performance parity with 41.8\% data reduction, outperforming InfoBatch's results at a lower 34.1\% pruning rate. This robustness is particularly evident in recognition tasks, as demonstrated on MJ+ST where SeTa preserves high accuracy (85.4\% on IC15) even with 40.4\% data reduction.
These results indicate that SeTa’s sliding window strategy successfully manages sample difficulty, preventing performance degradation even at high pruning rates.

\noindent{\textbf{Significant Pruning Capabilities (L3).}} At the highest pruning rates where slight performance degradation is acceptable, SeTa enables substantially higher pruning rates while maintaining competitive performance. This is particularly evident on MJ+ST, where SeTa achieves a 71.0\% data reduction while maintaining near-baseline performance (95.8\% vs 96.1\% on IIIT5k). This significantly exceeds InfoBatch's maximum achievable pruning rate of 50.3\%, suggesting that SeTa effectively removes redundant samples while maintaining the essential diversity of the training distribution, enabling efficient training without compromising performance.

These comprehensive results demonstrate SeTa's significant advantages over existing methods, particularly in its ability to achieve higher pruning rates while better preserving model performance. The consistent superior performance across diverse tasks and datasets suggests that SeTa's effectiveness stems from its fundamental ability to identify and retain essential training samples.

\begin{table}[h]
    \caption{WHU-MVS including 28K image-depth map samples for multi-view stereo task.}
    \label{tab:whu_mvs}
    \centering
    \footnotesize
    \setlength{\tabcolsep}{2pt}
    \begin{tabular}{cc|cccc}
        \toprule[1.3pt]
        Method & Pruned \% & MAE (m) $\downarrow$ & <3-interval (\%) $\uparrow$ & <0.6m (\%) $\uparrow$ \\
        \midrule[1.3pt]
        Ada-MVS & - & 0.1185 & 95.01 & 97.38 \\
        \hline

        \multirow{3}{*}{+SeTa}
            & 40.4 & 0.1184 & 95.28 & 97.44 \\
            & 52.4 & 0.1269 & 94.67 & 97.29 \\
            & 58.1 & 0.1357 & 94.24 & 97.08 \\
        \bottomrule[1.3pt]
    \end{tabular}
\end{table}
\begin{table}[h]
    \caption{CVACT including 35K image samples for cross-view geo-localization.}
    \label{tab:geo}
    \centering
    \footnotesize
    \begin{tabular}{cc|cccc}
        \toprule[1.3pt]
        Method & Pruned \% & R@1 & R@5 & R@10 & R@1\% \\
        \midrule[1.3pt]
        GeoDTR & - & 86.21 & 95.44 & 96.72  & 98.77 \\
        \hline

        \multirow{3}{*}{+SeTa}
            & 33.6 & 86.24 & 95.45 & 96.73 & 98.77 \\
            & 46.6 & 85.69 & 95.42 & 96.70 & 98.69 \\        
            & 56.5 & 85.25 & 95.17 & 96.52 & 98.66 \\        
        \bottomrule[1.3pt]
    \end{tabular}
\end{table}

\begin{table}[h]
    \caption{Alpaca including 52K instruction for tuning LLaMA-7B.}
    \label{tab:llm}
    \centering
    \footnotesize
    \setlength{\tabcolsep}{3pt}
    \begin{tabular}{cc|ccccc}
        \toprule[1.3pt]
        Method & Pruned \% & BBH & DROP & MMLU & Human-Eval & Avg. \\
        \midrule[1.3pt]
        DQ (2\%)~\cite{zhou2023dataset} & - & 32.9 & 27.6 & 36.6  & 8.5 & 26.3 \\
        \hline

        \multirow{3}{*}{+SeTa}
            & 25.9 & 32.4 & 27.4 & 35.8 & 13.4 & 27.3 \\
            & 40.3 & 32.3 & 27.0 & 35.4 & 11.6 & 26.6 \\        
            & 52.5 & 32.3 & 27.3 & 34.8 & 9.8 & 26.1 \\        
        \bottomrule[1.3pt]
    \end{tabular}
\end{table}

\begin{table*}[h]
    \caption{Cross-architecture generalization experiments on ImageNet-1K and CIFAR100 with CNNs, Transformers, and Mambas.}
    \label{tab:arch}
    \centering
    \footnotesize
    \begin{tabular}{c|ccccccc|cc}
        \toprule[1.3pt]
        \multirow{3}{*}{Method} & \multicolumn{7}{c|}{ImageNet-1K} & \multicolumn{2}{c}{\multirow{2}{*}{CIFAR100}} \\
        ~ & \multicolumn{4}{c}{CNN} & \multicolumn{2}{c}{Transformer} & Mamba & ~ & ~\\
        \cmidrule(lr){2-5} \cmidrule(lr){6-7}  \cmidrule(lr){8-8}
        ~ & R18 & R50 & MobileNetV3 & EfficientNet & ViT & Swin & Vim & R18 & R50 \\
        \midrule[1.3pt]
        Full
        & 69.5 & 78.6 & 52.2 & 76.1
        & 73.3  & 80.0 & 75.7
        & 79.3 & 80.6 \\
        \hline

        \multirow{3}{*}{+SeTa}
            & 69.6 / 36.1 & 78.5 / 30.0 & 52.2 / 26.2 & 76.3 / 30.0
            & 73.3 / 24.6 & 80.0 / 46.4 & 75.7 / 30.0
            & 79.4 / 41.6 & 80.7 / 40.9 \\

            & 68.5 / 44.2 & 78.1 / 45.5 & 52.0 / 42.8 & 75.7 / 45.0
            & 72.1 / 38.3 & 79.3 / 55.5 & 74.9 / 44.8
            & 79.0 / 54.9 & 80.2 / 54.1 \\

            & 68.2 / 55.3 & 77.6 / 52.7 & 51.3 / 50.8 & 75.6 / 52.3
            & 72.0 / 44.8 & 79.2 / 63.2 & 74.1 / 55.8
            & 78.4 / 60.6 & 79.5 / 68.3 \\
        \bottomrule[1.3pt]
    \end{tabular}
\end{table*}

\begin{table}[h]
    \caption{CIRR including 22K image-text samples for composed image retrieval.}
    \label{tab:cir}
    \centering
    \footnotesize
    \setlength{\tabcolsep}{1pt}
    \begin{tabular}{cc|cccc|ccc|c}
        \toprule[1.3pt]
        \multirow{2}{*}{Method} & \multirow{2}{*}{Pruned \%} & \multicolumn{4}{c|}{Recall@K} & \multicolumn{3}{c|}{Recall$_{subset}$@K}  & Avg  \\
        ~ & ~ & K = 1 & K = 5 & K = 10 & K = 50 & K = 1 & K = 2 & K =3 & R5, R$_{sub}$1 \\
        \midrule[1.3pt]
        CaLa & - & 43.4 & 74.7 & 84.3  & 96.3 & 72.0 & 87.6 & 94.4 & 73.4 \\
        \hline

        \multirow{2}{*}{+SeTa}
            & 60.5
            & 43.5 & 74.8 & 84.8 & 96.3
            & 72.2 & 88.4 & 94.8 & 73.5   \\

            & 70.0
            & 42.0 & 73.9 & 84.5 & 96.2
            & 70.7 & 87.1 & 94.0 & 72.3
            \\
        \bottomrule[1.3pt]
    \end{tabular}
\end{table}

\begin{table}[h]
    \caption{COCO including 556K text samples for image caption.}
    \label{tab:compare_caption_coco}
    \centering
    \footnotesize
    \setlength{\tabcolsep}{3pt}

    \begin{tabular}{cc|cccc|cccc}
        \toprule[1.3pt]
        \multirow{2}{*}{Method} & \multirow{2}{*}{Pruned \%} & \multicolumn{4}{c|}{NoCaps Val (CIDEr)} & \multicolumn{4}{c}{COCO} \\
        ~ & ~ & In & Near & Out & Overall & B@4 & M & C & S \\
        \midrule[1.3pt]
        \multirow{1}{*}{ViECap} & - 
            & 58.4 & 63.1 & 65.3 & 65.2
            & 27.1 & 24.6 & 91.5 & 18.0
            \\
        
        \hline

        \multirow{3}{*}{+SeTa} & 30.7
            & 58.7    & 63.2    & 65.6    & 65.4
            & 27.6    & 24.7    & 92.8    & 18.1
            \\
            
            ~ & 41.2
            & 57.7    & 61.9    & 64.6    & 64.1
            & 27.0    & 24.4    & 91.7    & 17.7
            \\
            
            ~ & 51.9
            & 57.0    & 61.6    & 64.4    & 63.9
            & 27.0    & 24.3    & 90.6    & 17.7
            \\
        
        \bottomrule[1.3pt]
    \end{tabular}
\end{table}

\begin{table}[h]
    \caption{RefCOCO including 142K image-text samples for referring image segmentation.}
    \label{tab:ris}
    \centering
    \footnotesize
    \begin{tabular}{cc|ccccc}
        \toprule[1.3pt]
        Method & Pruned \% & P@0.5 & P@0.7 & P@0.9 & oIoU & mIoU \\
        \midrule[1.3pt]
        \multirow{1}{*}{CARIS} & - 
            & 86.7 & 79.3 & 40.4 & 74.5 & 76.7 \\

        \hline

        \multirow{3}{*}{+SeTa} & 25.2
            & 86.7 & 79.8 & 40.7 & 74.2 & 76.6
            \\
            
            ~ & 40.0
            & 86.3 & 78.9 & 40.0 & 73.6 & 76.1
            \\
            
            ~ & 50.2
            & 85.4 & 77.9 & 38.9 & 72.4 & 75.9
            \\
        
        \bottomrule[1.3pt]
    \end{tabular}
\end{table}

\subsection{Generalization Evaluation}
We further evaluate SeTa's generalization capability across diverse vision and language tasks, different model architectures, and various scales of datasets. The experimental results (Table~\ref{tab:whu_mvs},\ref{tab:geo},\ref{tab:llm},\ref{tab:arch},\ref{tab:cir},\ref{tab:compare_caption_coco},\ref{tab:ris}) demonstrate that SeTa exhibits strong generalization ability in three key aspects.

\noindent{\textbf{Task Generalization.}} SeTa demonstrates remarkable effectiveness across diverse vision and language tasks.

(1) For dense prediction tasks like multi-view stereo (Table~\ref{tab:whu_mvs}) and RefCOCO (Table~\ref{tab:ris}): SeTa achieves superior performance across both tasks. For multi-view stereo, SeTa achieves better MAE (0.1184 vs 0.1185) while pruning 40.4\% data. The improvement in <3-interval metric (95.28\% vs 95.01\%) suggests better handling of depth discontinuities. Even at higher pruning rates (52.4\%), performance degradation is minimal (MAE increase of only 0.0084). This indicates SeTa effectively identifies key samples that capture complex geometric relationships. Similarly, on RefCOCO, SeTa demonstrates improved performance (R@1: 86.24 vs 86.21) with 33.6\% data reduction, indicating its ability to preserve crucial semantic alignment information.

(2) For retrieval tasks: On cross-view geo-localization (Table~\ref{tab:geo}), SeTa maintains performance across all metrics (R@1: 86.24 vs 86.21) with 33.6\% data reduction. On composed image retrieval (Table~\ref{tab:cir}), SeTa achieves better results (R@5: 74.8 vs 74.7) even with substantial 60.5\% pruning. The consistent improvement in both tasks suggests SeTa preserves crucial semantic alignment information. Notably, performance remains stable even at higher recall thresholds (R@50), indicating robust feature learning.

(3) For generation tasks: On instruction tuning (Table~\ref{tab:llm}), SeTa enhances overall performance (Avg: 27.3 vs 26.3) while pruning 25.9\% data. Particularly strong improvements in Human-Eval (13.4 vs 8.5) suggest better preservation of code-related examples. On COCO caption generation, SeTa achieves significant gains (CIDEr: 92.8 vs 91.5) with 30.7\% data reduction. The improvement spans both in-domain and out-of-domain (NoCaps), indicating better generalization.

SeTa's effectiveness is particularly pronounced in tasks requiring fine-grained understanding (dense prediction, retrieval). The method maintains or improves performance even with substantial data reduction (40-60\%).
This comprehensive evaluation demonstrates SeTa's task-agnostic nature and its ability to identify essential training samples regardless of the underlying task complexity or data modality.

\noindent{\textbf{Architecture Generalization.}} Table~\ref{tab:arch} demonstrate SeTa's strong generalization capability across diverse architectures:

(1) For CNN-based models, SeTa shows remarkable efficiency across different architectures:
ResNet-18 maintains its 69.5\% accuracy while achieving 36.1\% data reduction.
The deeper ResNet-50 preserves its higher 78.6\% accuracy with 30.0\% data pruning.
Notably, efficient architectures like MobileNetV3 and EfficientNet, which are specifically designed for resource-constrained environments, also benefit from SeTa:
MobileNetV3 maintains its 52.2\% accuracy with 26.2\% data reduction, while EfficientNet preserves its 76.1\% accuracy with 30.0\% pruning.
This demonstrates that SeTa can further optimize already-efficient architectures, making them even more suitable for resource-limited scenarios.

(2) For Transformer architectures:
SeTa demonstrates consistent effectiveness across different variants:
Swin maintains its strong 80.0\% accuracy while achieving substantial 46.4\% data reduction.
ViT preserves its 73.3\% accuracy with 24.6\% pruning, showing more conservative but still significant data efficiency.
The performance preservation at higher pruning rates is particularly notable:
Swin maintains 79.3\% accuracy even with 55.5\% data reduction.

(3) For Vision Mamba architectures:
Vim maintains its 75.7\% accuracy while pruning 30.0\% of data. The model exhibits similar pruning tolerance to ResNet-50 and EfficientNet (all at 30.0\%), suggesting that despite their architectural differences, these networks have comparable data efficiency characteristics. Even at higher pruning rates (44.8\%), Vim maintains relatively strong performance (74.9\%), demonstrating its robustness to data reduction. This resilience is particularly noteworthy given Mamba's recent emergence and distinct architecture from both CNNs and Transformers.

SeTa exhibits cross-architecture consistency, ensuring that all architectures preserve their baseline performance.
The pruning rates (26.2-46.4\%) are consistent across architectural families, with efficiency-oriented models showing slightly more conservative pruning rates.
This architecture-agnostic behavior suggests SeTa captures fundamental data importance rather than model-specific features.

\noindent{\textbf{Dataset Scale Generalization.}} SeTa demonstrates robust generalization across datasets of vastly different scales:

(1) For large-scale datasets (>1M samples):
On ImageNet (1.28M samples), SeTa maintains competitive performance while achieving substantial data reduction across different architectures (30.0-46.4\%). Even at higher pruning rates (>50\%), performance degradation is gradual rather than catastrophic, with accuracy drops of only 1-2\% points. This suggests SeTa effectively identifies essential training samples even in large-scale settings.

(2) For medium-scale datasets (100K-1M samples):
On COCO (556K samples), SeTa not only maintains but improves performance (CIDEr: 92.8 vs 91.5) with 30.7\% data pruning. The improvements span both in-domain and out-of-domain evaluation, with gains in NoCaps Overall (65.4 vs 65.2). This indicates SeTa's ability to identify and retain informative samples that contribute to model generalization.

(3) For small-scale datasets (<100K samples):
On CIRR (22K samples), SeTa achieves superior results (R@5: 74.8 vs 74.7) even with substantial data reduction (60.5\%). Notably, performance remains stable across different evaluation metrics (R@1, R@5, R@10), suggesting that SeTa effectively preserves the dataset's diversity despite its small size. This is particularly impressive given that smaller datasets typically offer less redundancy to exploit.

SeTa achieves scale-agnostic data efficiency, consistently retaining critical samples with 30-50\% data compression while maintaining state-of-the-art performance across varying data scales, demonstrating robust generalization capability and practical utility for real-world deployment.

In summary, SeTa is a versatile data selection method that can be effectively applied to diverse tasks, models, and datasets, including both synthetic and real datasets, without requiring task-specific adjustments.

\subsection{Ablation Study}

\noindent{\textbf{Components.}} We conduct ablation studies to analyze the effectiveness of components in SeTa as shown in Table~\ref{tab:comp}.

(1) \textit{Dynamic vs Static Sampling:} We first compare dynamic sampling (randomly sampling from the full dataset each epoch) and static sampling (sampling once before training) as baselines. SeTa outperforms both approaches across all metrics, with particularly notable improvements in Out-domain (71.4 vs 69.8/69.4) and Overall (71.5 vs 69.9/69.4) performance on NoCaps. This demonstrates the effectiveness of SeTa's sliding window strategy over simple random.

(2) \textit{Component Analysis:} We ablate three key components of SeTa:
Hard-to-easy (h2e) ordering achieves lower performance (69.6 Overall) compared to SeTa's easy-to-hard approach (71.5 Overall), suggesting that progressive difficulty is beneficial for model learning.
Continuous scheduling, which does not directly set $s$ to zero, shows strong performance (70.7 Overall) but still falls short of the full SeTa framework, indicating that maintaining consistent data patterns throughout training is important.
Annealing strategy, similar to InfoBatch, which uses full data for training in the last few epochs to eliminate bias, shows good performance (70.3 Overall) but does not reach the same level as SeTa's partial annealing, while also reducing the amount of data.

(3) \textit{Efficiency-Performance Trade-off:} Notably, SeTa achieves superior performance (71.5 Overall) compared to training on the full dataset (70.5 Overall) while using only 68.3\% of the data. This indicates that SeTa's data selection strategy not only reduces computational cost but also helps filter out potentially noisy or less informative samples that might hinder model learning.

These results demonstrate that SeTa's success stems from the synergistic combination of its components, with each contributing to its effectiveness in data-efficient training.

\begin{table}[h]
    \caption{Effect of different components.}
    \label{tab:comp}
    \centering
    \footnotesize
    \setlength{\tabcolsep}{5pt}
    \begin{tabular}{ccc|c|cccc}
        \toprule[1.3pt]
        ~ & \multirow{2}{*}{Operation} & \multirow{2}{*}{Pruned \%} & COCO & \multicolumn{4}{c}{NoCaps Val (CIDEr)} \\
        ~ & ~ & ~ & Test & In & Near & Out & Overall \\
        \midrule[1.3pt]
        \multirow{2}{*}{\rotatebox[origin=c]{270}{Base}} & dynamic & 31.7 & 94.7 & 62.3 & 67.9 & 69.9 & 69.8 \\
        ~ & static & 31.7 & 94.6 & 61.5 & 67.5 & 69.4 & 69.4 \\
        \addlinespace[2pt]
        \hline
        \addlinespace[2pt]
        \multirow{3}{*}{\rotatebox[origin=c]{270}{Comp.}} & h2e & 27.0 & 94.8 & 62.0 & 67.3 & 70.3 & 69.6 \\
        ~ & continuous & 32.8 & 95.1 & 63.1 & 68.2 & 71.3 & 70.7 \\
        ~ & anneal & 27.2 & 95.2 & 62.6 & 68.1 & 71.3 & 70.3 \\
        \addlinespace[2pt]
        \hline
        \addlinespace[2pt]
        ~ & Full & - & 95.2 & 63.2 & 68.8 & 70.2 & 70.5 \\
        ~ & SeTa & 31.7 & \textbf{95.3} & \textbf{63.4} & \textbf{69.7} & \textbf{71.4} & \textbf{71.5} \\

        \bottomrule[1.3pt]
    \end{tabular}
\end{table}

\begin{figure*}[ht]
    \centering
    \begin{subfigure}[t]{0.33\textwidth}
        \includegraphics[width=\linewidth]{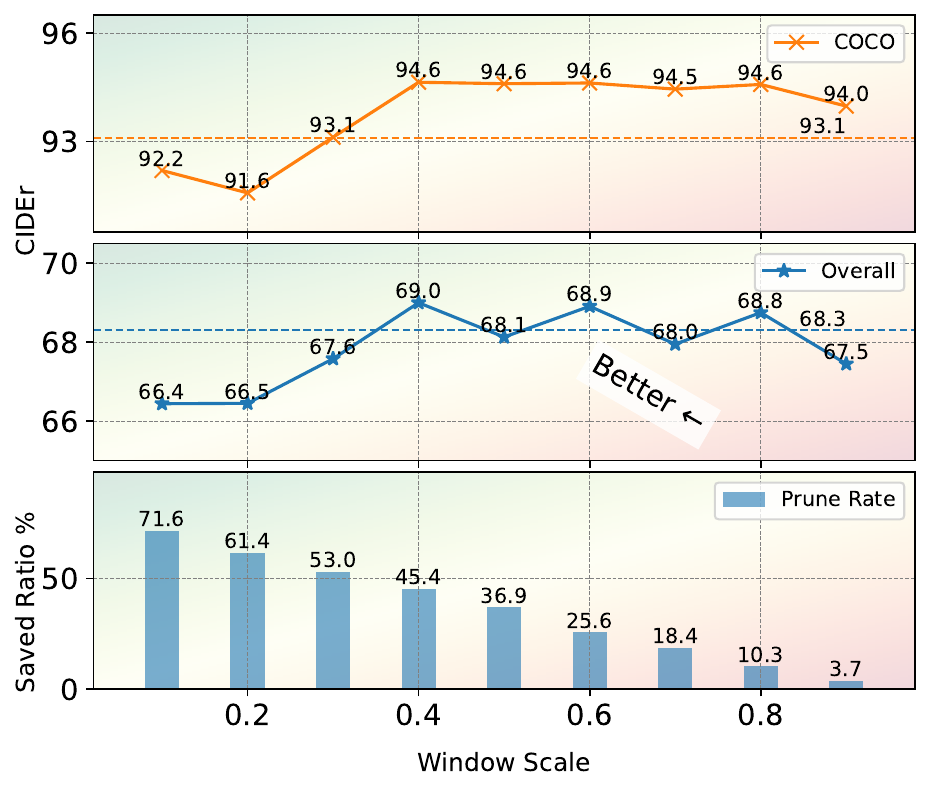}
        \caption{Effect of window scale $\alpha$ on performance.}
        \label{fig:ab-ws}
    \end{subfigure}
    \begin{subfigure}[t]{0.33\textwidth}
        \includegraphics[width=\linewidth]{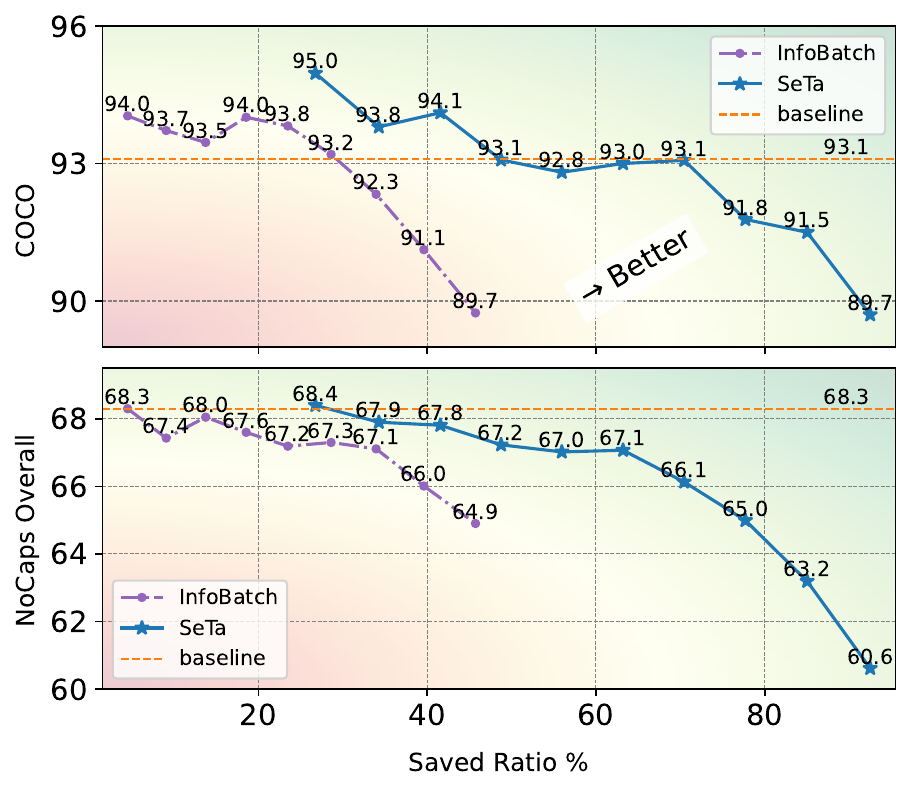}
        \caption{Methods comparison across ratios $r$.}
        \label{fig:ab-hp-comp}
    \end{subfigure}
    \begin{subfigure}[t]{0.325\textwidth}
        \includegraphics[width=\linewidth]{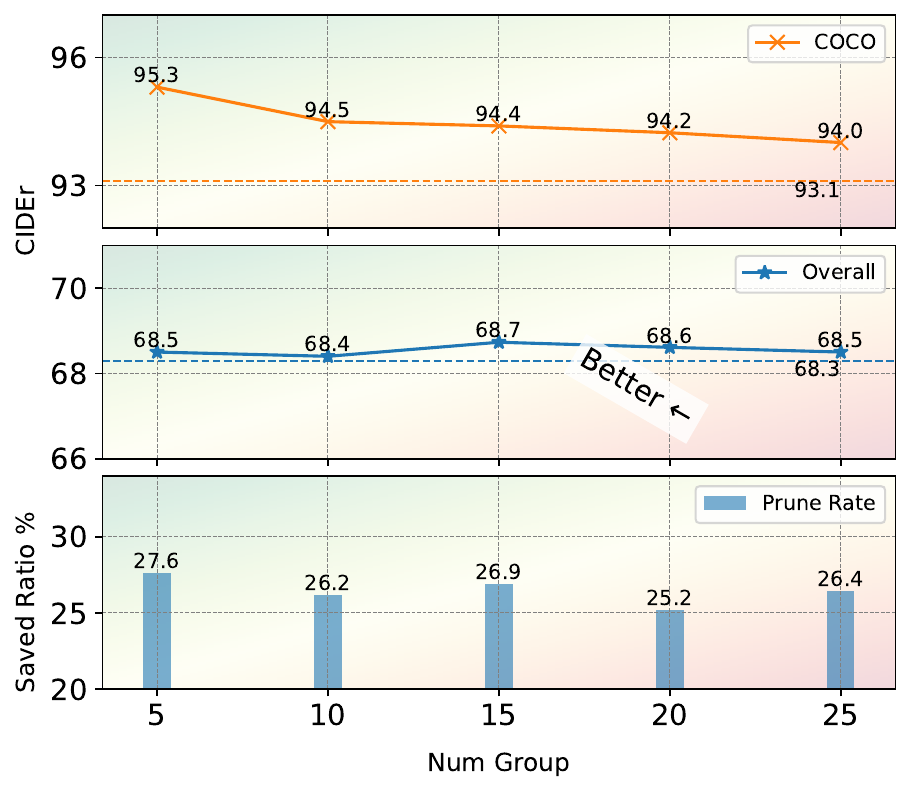}
        \caption{Effect of group num $k$ on performance.}
        \label{fig:ab-hp-ng}
    \end{subfigure}
    \vspace{-2pt}
    \caption{Hyperparameter ablation study on ToCa-1M}
    \label{fig:ab}
\end{figure*}

\noindent{\textbf{Hyperparameter.}}
We conduct extensive experiments to analyze the sensitivity and impact of key hyperparameters in SeTa, with results shown in Figures~\ref{fig:ab}.

(1) \textit{Window Scale:} As shown in Figure~\ref{fig:ab-ws}, we observe several key trends:
The optimal window size exhibits a sweet spot around 0.4-0.6 of the dataset size, balancing between sufficient context and computational efficiency.
Performance degrades sharply with very small windows (<0.2) due to insufficient learning context.
Larger windows (>0.7) show diminishing returns while increasing computational cost.

(2) \textit{Pruning Strategy Comparison:} Figure~\ref{fig:ab-hp-comp} reveals important insights about different pruning approaches:
SeTa maintains consistently higher performance across all pruning ratios compared to SOTA methods InfoBatch.
The performance gap between SeTa and InfoBatch widens as the pruning ratio increases, demonstrating its superior robustness.
The gradual slope of SeTa's performance curve indicates graceful degradation under aggressive pruning.

(3) \textit{Group Number:} Figure~\ref{fig:ab-hp-ng} shows that SeTa is relatively robust to the choice of group numbers:
Performance remains stable across a wide range of group numbers (5-15), with minor fluctuations.
While there is a slight degradation, the impact on overall performance is minimal.

These findings provide practical guidelines for hyperparameter selection:
Window scale of 0.4-0.6 offers the best performance-efficiency trade-off.
Downsampling ratios up to 0.3 can be used with minimal performance degradation.
Group numbers can be flexibly chosen within 5-15 without significant performance impact.

\begin{figure}[ht]
    \centering
    \includegraphics[width=\linewidth]{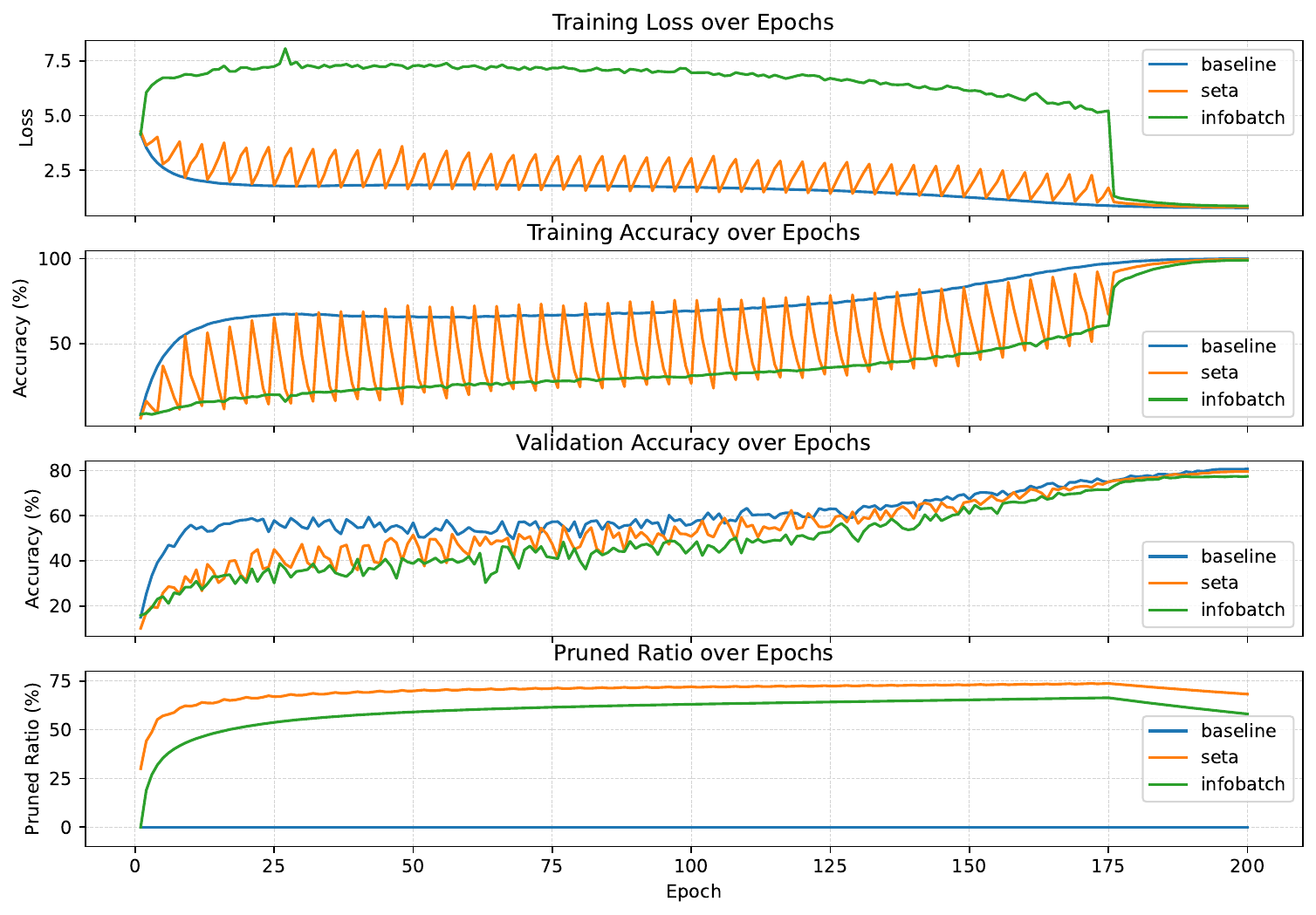}
    \vspace{-15pt}
    \caption{Training process comparison of full dataset (baseline), SeTa and InfoBatch on CIFAR100.}
    \label{fig:work}
    \vspace{-15pt}
\end{figure}

\noindent{\textbf{Validation.}}
We further validate the effectiveness of SeTa by analyzing its training dynamics across four key metrics - training loss, training accuracy, validation accuracy, and data saving ratio on CIFAR100, as shown in Figure~\ref{fig:work}.

(1) \textit{Oscillatory Learning Pattern:} SeTa exhibits characteristic periodic oscillations between the performance levels of full dataset training and InfoBatch. These controlled fluctuations, driven by the sliding window selection strategy, serve a crucial role in preventing the model from converging to local optima while maintaining higher data efficiency (shown by the consistently higher data saving ratio).

(2) \textit{Efficient Learning Dynamics:} Despite using significantly less data (shown by the higher data saving curve), SeTa maintains competitive performance with full dataset training. The periodic oscillations in both training and validation accuracy suggest that the sliding window mechanism effectively refreshes the learning signal, allowing the model to explore different aspects of the data distribution while maintaining stable overall performance.

(3) \textit{Robust Convergence:} The validation accuracy curve demonstrates that SeTa achieves comparable final performance to full dataset training, while consistently outperforming InfoBatch. This indicates that the oscillatory learning pattern induced by sliding window selection not only improves training efficiency but also leads to robust convergence.

These observations validate that SeTa's superior performance stems from its ability to leverage controlled performance oscillations through sliding window selection, balancing between data efficiency and learning stability.

\section{Conclusion}
We present SeTa, a simple yet effective dynamic pruning framework for efficient large-scale training. Through extensive experiments, we demonstrated that SeTa reduces training costs by 30-50\% while preserving or even enhancing model performance. Its sliding window strategy adaptively focuses on informative samples, filtering out redundant and inefficient ones. SeTa is model-agnostic and integrates seamlessly into existing training pipelines without architectural modifications. Experiments on various real-world datasets further validate its strong generalization across architectures and tasks, offering a practical solution to the growing computational challenges in deep learning.

\section*{Acknowledgements}
This work was supported by the Natural Science Foundation of China under Grant 62471394, U21B2041, and 62306241.

\bibliographystyle{abbrv}
{
	\small
	\bibliography{main}
}

\clearpage

\end{document}